\documentclass[conference]{IEEEtran}
\IEEEoverridecommandlockouts

\usepackage{graphicx}
\usepackage{amsmath}
\usepackage{amssymb}
\usepackage{booktabs}
\usepackage{balance}

\usepackage[pagebackref,breaklinks,colorlinks]{hyperref}

\usepackage[capitalize]{cleveref}
\crefname{section}{Sec.}{Secs.}
\Crefname{section}{Section}{Sections}
\Crefname{table}{Table}{Tables}
\crefname{table}{Tab.}{Tabs.}

\def\cvprPaperID{17064}
\def\confName{CVPR}
\def\confYear{2025}

\begin{document}

\title{ContRail: A Framework for Realistic Railway Image Synthesis using ControlNet}
\author{
    \IEEEauthorblockN{Andrei-Robert Alexandrescu\IEEEauthorrefmark{1},  R\u{a}zvan-Gabriel Petec\IEEEauthorrefmark{1}, Alexandru Manole\IEEEauthorrefmark{1} and Laura-Silvia Dio\c{s}an\IEEEauthorrefmark{1}}
    \IEEEauthorblockA{\IEEEauthorrefmark{1}Department of Computer Science, Babe\c{s}-Bolyai University, Cluj-Napoca, Romania\\
    Email:  andrei.robert.alexandrescu@gmail.com, \{razvan.petec, alexandru.manole, laura.diosan\}@ubbcluj.ro}
}

\maketitle

\begin{abstract}
Deep Learning became an ubiquitous paradigm due to its extraordinary effectiveness and applicability in numerous domains. However, the approach suffers from the high demand of data required to achieve the potential of this type of model. An ever-increasing sub-field of Artificial Intelligence, Image Synthesis, aims to address this limitation through the design of intelligent models capable of creating original and realistic images, endeavour which could drastically reduce the need for real data. The Stable Diffusion generation paradigm recently propelled state-of-the-art approaches to exceed all previous benchmarks. In this work, we propose the ContRail framework based on the novel Stable Diffusion model ControlNet, which we empower through a multi-modal conditioning method. We experiment with the task of synthetic railway image generation, where we improve the performance in rail-specific tasks, such as rail semantic segmentation by enriching the dataset with realistic synthetic images.

\end{abstract}

\section{Introduction}
\label{sec:intro}

Recent years have seen a significant increase in the size of artificial neural network models, with architectures reaching billions of parameters \cite{bansal2022systematic}. Training such models is not trivial, given the immense time and data required to train them properly. Therefore, data collection and annotation represent crucial components of the supervised learning process, especially for large models.

In some particular cases, data collection and annotation can be expensive \cite{whang2023data}. For instance, in the field of vision understanding for autonomous driving, image segmentation techniques are used to comprehend the surrounding environment. These models require highly detailed ground-truth maps of the environment, assigning each pixel to a specific class from a set of predefined classes. This process is laborious and costly when using third-party services.

Instead of gathering and labeling real images, researchers employ image generation techniques to create synthetic images as viable alternatives. This approach expands the possibilities for model training, as it theoretically provides an infinite amount of data. However, while this is advantageous quantitatively, the quality and realism of the generated images are crucial. The distribution from which these synthetic images are sampled must closely resemble real-world data to ensure an effective training process \cite{eigenschink2023deep}.

In this article, we study the task of Railway Image Generation using ControlNets \cite{zhang2023adding} to generate realistic images taken from the ego-view of the train. We evaluate them using the Fr\'{e}chet Inception Distance (FID) \cite{heusel2017gans}, which is a metric that aims to quantify the realism of images. The images are then used to improve the scores of a model that performs Semantic Segmentation on the rails \cite{alexandrescu2022dynamic}. We show an increase in performance when enhancing the training corpus with synthetic images. We also experiment with various bimodal image representations and discuss their results.

Even though there are various image synthesis models (e.g., ControlNet \cite{zhang2023adding}, T2I Adapter \cite{mou2024t2i}, SCEdit \cite{jiang2024scedit}), we select ControlNet because of its more precise control. The condition guides the generation process step by step, being integrated directly into the diffusion model. Furthermore, ControlNet can produce higher-quality images with more intricate details, of course at a high cost in terms of the model size and the processing time. While this trade-off cannot be ignored, we aim to show how synthetic images can help train high-performing segmentation models.

Working with single image representations has limitations when compared to a more robust representation that concatenates multiple modalities, such as segmentation masks and Canny edges. Therefore, we combine the previously mentioned representations into one in more than one ways and study the model's behavior to input variance. 

This research aims to provide meaningful answers to the following research questions:


\begin{itemize}
\item Can we improve the performance of scene understanding intelligent models in the data-scarce rail domain using synthetic images?
\item What is the ideal conditional representation for railway image generation, measured through visual inspection and FID minimization?
\end{itemize}


To answer to these research questions, we propose \emph{ContRail}, a ControlNet-based approach, tailored for the railway scene understanding domain, which exploits a novel representation of input conditions combining a semantic segmentation mask and an edge image. As we do not have captions for our input images, we experiment with three types of prompts (empty, fixed and generated using BLIP2 \cite{li2023blip}). We perform an ablation study to measure the impact of each proposed component and showcase the impressive quality of the synthesized images by training a semantic segmentation model using the generated data and testing it on real samples.

The remainder of the paper is structured as follows: Section \ref{sec:concepts} introduces related concepts, Section \ref{sec:relWork} showcases similar applications, Section \ref{sec:methodology} describes the methodology used in our work to generate realistic images, Section \ref{sec:results} showcases visual and numerical results, and Section \ref{sec:conclusion} provides a conclusion to the article, as well as future considerations.

\section{Concepts}
\label{sec:concepts}

\subsection{Semantic Segmentation}
\label{sec:semanticSegmentation}

Semantic Segmentation is a computer vision task which consists of assigning class labels to each pixel from a given image. The output of this task is much more granular when compared to other Image Recognition problems including image classification and object detection, as the label is attached to each individual pixel. Semantic Segmentation can be either binary, where the pixels are classified as part of the class of interest or as background, or multi-clas, where each pixel has multiple values attached to it, each representing the probability of the pixel being part of a certain class.

One popular metric for this task is Intersection over Union (IoU) \cite{jaccard1912distribution}, described as:

\begin{equation}
    mIoU = \frac{|Y \cap Y_p|}{|Y \cup Y_p|},
\end{equation}

\noindent
where $Y$ and $Y_p$ are the ground-truth and predicted masks with the segmentation results. The intersection of similar pixels is divided with their union, on average for all classes.

\subsection{Latent Diffusion Models}
\label{sec:stableDiffusion}

Diffusion Models \cite{sohl2015deep} are unsupervised machine learning models used to estimate an assumed probability distribution of some observed data. This kind of models are often called Generative Models. The idea of these models is to slowly destroy structure in the observed data distribution through a process called {\it diffusion}, then to learn a {\it reverse diffusion process} that restores the structure in that data.

The diffusion process starts from an observed data point $x_0$, then some Gaussian noise is combined with the last input using a noise schedule parameter $\beta_{1..\mathcal{T}} \in (0,1)^\mathcal{T}$, where $\mathcal{T}$ is the maximum number of diffusions added to some input, resulting in a $t$-times (usually called the {\it timestep} of the diffusion) diffused data point $x_t = \sqrt{1 - \beta_t} \cdot x_{t-1} + \sqrt{\beta_t} \cdot \epsilon_t$, where $\epsilon_t \sim \mathcal{N}(0, 1)$.

The reverse diffusion process usually consists of predicting the noise $\epsilon$ added at timestep $t$ to some input data point $x_t$. It is sufficient to find the noise $\epsilon$ instead of the data point at the previous timestep $x_{t-1}$. The noise is used to compute the original data point as $x_{t-1} = \frac{x_t - \sqrt{\beta_t} \cdot \epsilon_t}{\sqrt{1 - \beta_t}}$. 

The objective of the machine learning model $\epsilon_\theta$ that computes the reverse diffusion process is to predict correctly the noise added from timestep $t - 1$ to timestep $t$ given the timestep $t$ and the noisy image $x_t$. The loss function is:

$$
L_{\text{DM}} = \mathbb{E}_{x_0, \epsilon \in \mathcal{N}(0, 1), t} \left[ \Vert \epsilon - \epsilon_\theta(x_t, t) \Vert^2_2 \right].
$$

The generation pipeline starts from noise $\epsilon_i \in \mathcal{N}(0, 1)$, diffused 5 times: $x_{\text{new}} = \epsilon_\theta(...(\epsilon_\theta(\epsilon_i, t), t - 1), ..., 2), 1)$.


Using the mean square error to compare each pixel leads to small differences in the pictures, such as a small shift of an object, a wrong color of an object, and other. Since this leads to large loss values, other metrics must be used to measure the differences between the pictures. One such metric is the perceptual loss \cite{johnson2016perceptual}, which uses a pre-trained convolutional neural network that is already proficient in extracting high-level features from the target picture and the predicted one to compute the mean square error between the features. This approach is computationally difficult, leading to a higher problem complexity.

To avoid the computation of perceptual loss, Latent Diffusion Models (LDMs) were introduced by Rombach et al. \cite{rombach2022high}. The main idea is to diffuse and reverse the diffusion process in the feature space of the inputs (instead of training it directly on the image space) and have a mechanism that can construct the original input from the features. The input space is often called the {\it latent space}, which is a lower-dimensionality space, leading to faster training of the network responsible for the reverse diffusion process.

Stable Diffusion \cite{runway2022stable} is a model that proves the capabilities of LDMs. It was trained on the LAION-5B dataset \cite{schuhmann2022laion}, which is a large-scale public image-text dataset of over 5.8 billion pairs of image and text annotations, making it one of the most robust models at the moment.

\subsection{ControlNet}
\label{sec:controlNet}

Although Stable Diffusion \cite{runway2022stable} is a very robust and complex model, it is not good enough for specific tasks, such as generating images in a specific environment, where the probability distribution of the data is most likely different from the one the model was trained on. In case we  want to make the model understand some of the new probability distribution, but also to keep some of the knowledge about the old distribution (such as how to handle diverse text embeddings, which may not be present in the new training data), fine-tuning the trained LDM directly tends to make it generate more replications, especially when it's being trained with a smaller numbers of samples \cite{perera2023analyzing}. Moreover, there is evidence that directly fine-tuning an LDM can lead to overfitting, mode collapse, and catastrophic forgetting \cite{zhang2023adding}.

ControlNet \cite{zhang2023adding} was proposed to solve this problem. It is composed of a HyperNetwork \cite{ha2016hypernetworks}, which is a smaller network trained to influence the weights of a larger network, which tries to solve the problem mentioned before, over some LDM. The purpose of this network is to add some conditional control (segmentation masks, Canny edges, human poses, etc.) to a locked production-ready LDM, without modifying its weights. This kind of control may be added directly using the cross-attention mechanism of the U-Net in LDMs, however, as we mentioned previously, it is problematic to do so, especially in case of smaller datasets.

Having a pre-trained neural block from the original LDM, a ControlNet block could be added by locking (freezing) the parameters of the original block and simultaneously cloning the block to a trainable copy. The trainable copy will receive an external conditioning vector $c$ as input. When such structure is applied to large models like Stable Diffusion, the locked parameters preserve the original model trained on billions of images, while the trainable copy will reuse the original model to establish a deep, robust, and strong backbone for handling diverse input conditions \cite{zhang2023adding}. The neural architecture is connected with ”zero convolutions” (zero-initialized 1x1 convolution layers) that progressively grow the parameters from zero and ensure that no harmful noise could affect the fine-tuning \cite{zhang2023adding}.

Because the LDM does the diffusion and the reverse diffusion operations in the latent space of its Autoencoder, ControlNet transforms the external conditioning vector through a small convolutional network $E(\cdot)$ to match the shape of the latent space of the LDM, then the control and the original latent value are added before passing through the rest of the ControlNet \cite{zhang2023adding}. Usually, the conditioning image matches the shape of the expected output, so the set that the ControlNet is trained on is $S = \{(x_i, c_i) | c_i - \text{the conditioning image}, x_i - \text{the image that the ControlNet should generate}\}$. Due to $x_i$ having the same shape as $c_i$, the transformations done by $E(\cdot)$ on the conditioning images should produce a compatible shaped output as the transformations done by $\mathcal{E}(\cdot)$.

\subsection{BLIP-2}
\label{sec:BLIP-2}

Proposed by Li et al. \cite{li2023blip}, BLIP-2 (Bootstrapped Language-Image Pre-training) is a model capable of understanding visual and textual information and generating text. It uses a pre-trained visual image transformer to encode the features of the images, which are then passed through a proposed network called Q-Former, a lightweight transformer which employs a set of learnable query vectors to extract visual features from the frozen image encoder and act as a bridge between the image and language models. The output is then passed through a pre-trained large language model (LLM), which offers a strong language generation to complete the output based on the previous network context. BLIP-2 achieves state-of-the-art results on various tasks, including image labeling and description, answering queries about images, and finding description-matching images.


\section{Related Work}
\label{sec:relWork}

In the context of autonomous driving for trains, semantic segmentation is a crucial task for understanding visual elements from the ego-view of the train. A large amount of labeled data is required to train robust for semantic segmentation. However, the available data in the rails domain is limited. One possible solution to this challenge is data augmentation, which also reduces the chance of overfitting. Simple operations such as random rotation, crop, or noise injection are commonly used, however, these methods are not sufficient when dealing with complex data \cite{kebaili2023deep}. Therefore, generative models can be used to generate images and enrich the training dataset. We discuss in the following subsections the impact of various generative models, including Generative Adversarial Networks (\ref{subsec:gan_augmentation}) and Diffusion Models (\ref{subsec:ldm_augmentation}), on augmenting different datasets.

\subsection{Data Augmentation using Generative Adversarial Networks}
\label{subsec:gan_augmentation}

In the field of medical imaging, particularly in the context of liver lesion classification, a small number of labeled data creates a significant challenge. To address this, Frid et al. \cite{frid2018gan} used Generative Adversarial Networks (GANs) \cite{goodfellow2014generative} to augment the dataset with synthetic images of liver lesions. GANs can implicitly learn the distribution of real images, and they are used to generate new samples that enhance the training set. They compared the effect of training a Convolutional Neural Network on the raw dataset, on the dataset augmented with classic transformations, and on the dataset augmented with both the GAN and the classic transformations. They measured the total accuracy of the Convolutional Neural Network trained on those datasets and obtained 57\% when training on the raw dataset, 78.6\% when training on the dataset augmented with classic techniques, and 85.7\% when training on the dataset augmented with both classic techniques and GAN inferences. The quality of the GAN generations was measured by asking two expert radiologists to classify real images, respectively synthetic generated images. They both obtained similar results when classifying each type of image: expert 1 managed to classify correctly 78\% of the real images and 77.5\% of the synthetic images, while expert 2 managed to classify correctly 69.2\% of the real images and 69.2\% of the synthetic images. This indicates the validity of the lesion generation process.

\subsection{Data Augmentation using Diffusion Models}
\label{subsec:ldm_augmentation}

Still in the medical field, particularly for brain MRI segmentation, Fernandez et al. \cite{fernandez2022can} proposed the brainSPADE pipeline to enrich the datasets, using Latent Diffusion Models (LDM) \cite{sohl2015deep} to generate new segmentation masks, and the SPADE architecture \cite{park2019semantic} to generate high quality images from semantic maps. To generate different types of segmentation masks, two different LDMs were trained on healthy and tumour-affected semantic label slices. To verify the ability to learn to segment healthy regions using synthetic data, they trained 2 models: $R_\text{iod}$, trained with real data (7008 real samples), and $S_\text{iod}$, trained with synthetic data (20000 generated samples).

For the first experiment, they validated the models on the test dataset from the same distribution the $R_\text{iod}$ model was trained on, and computed different Dice scores \cite{dice1945measures}. In this test, the $R_\text{iod}$ model with a CSF Dice score of $0.958 \pm 0.008$ outperformed the $S_\text{iod}$ model with a CSF Dice score of $0.919 \pm 0.023$, but the result of $S_\text{iod}$ was comparable to the other performances achieved in the literature.

For the second experiment, they collected out of distribution (OoD) images, which are samples taken from various datasets with different styles. These out of distribution images were then split in near out of distribution (n -- OoD) samples (which have similar features between them) and far out of distribution (f -- OoD) samples (with more varying features). Then, they trained the models $R_\text{n -- OoD}$, respectively $S_\text{n -- OoD}$ with real images from the n-OoD samples, respectively images generated to match the distribution of the n-OoD samples, and the models $R_\text{f -- OoD}$, respectively $S_\text{f -- OoD}$, with meanings similar to the ones before, but using the f -- OoD samples. For the n -- OoD samples, the $S_\text{n -- OoD}$ model (with a CSF Dice score of $0.914 \pm 0.022$) outperformed the $R_\text{n -- OoD}$ model (with a CSF Dice score of $0.841 \pm 0.017$), and for the f -- OoD samples, the $S_\text{f -- OoD}$ model (with a CSF Dice score of $0.830 \pm 0.050$) also outperformed the $R_\text{f -- OoD}$ model (with a CSF Dice score of $0.792 \pm 0.034$). This showed that the brainSPADE pipeline has more potential for domain adaptation, being able to replicate to some extent the style of an unseen dataset.

In the field of autonomous driving, Khullar et al \cite{khullar2023synthetic} generated synthetic images with rare scenarios containing emergency vehicles, and showed an improvement in object detector models trained on them. They used a pre-trained diffusion model together with CLIP to condition the generation process \cite{radford2021learning}.

Carlson et al. \cite{carlson2023generation} presents a technique for traffic sign generation using diffusion models. They evaluated the synthetic images on a traffic sign classification task and showed improvements when using generated data for training.

All of these works support the impact of Diffusion Models on improving models trained for various tasks. While the medical domain applications are more frequent due to the scarcity of annotated samples, the field of autonomous driving can also highly benefit from the usage of synthetic images \cite{eigenschink2023deep}.

\section{Proposed Approach}
\label{sec:methodology}

\begin{figure*}[hbt!]
\centering
\centerline{\includegraphics[scale=0.25]{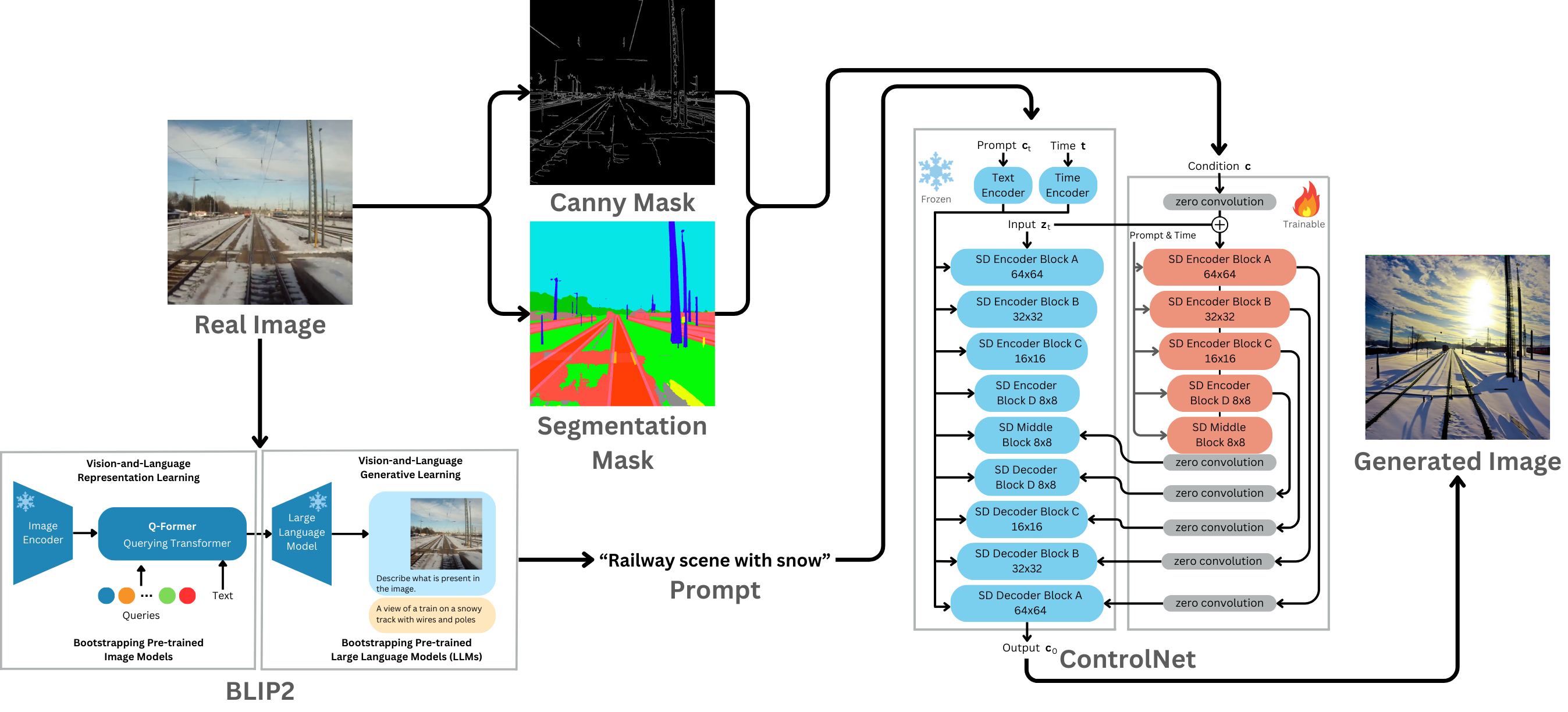}}
	\caption{ Overview of the proposed railway scene generation pipeline using ControlNet \cite{zhang2023adding} and BLIP-2 \cite{li2023blip}. RailSem19 provides real images and their semantic segmentation masks. Starting from the real image we can obtain the edges using the Canny algorithm. The segmentation mask is combined with the edge image as showed in Fig. \ref{fig:conditional_representation}. The original image is also used as input for the BLIP2 model which obtains its textual description, which is used for prompting purposes. The text and the combined conditional representation is fed into the ControlNet architecture, resulting in a realistic synthetic image.
 }
	\label{fig:apprach}
\end{figure*}

Our work aims to generate realistic railway images to compensate for the lack of annotated data from this domain. 
We propose an authentic railway scene image synthesis pipeline based on the ControlNet architecture \cite{zhang2023adding}. 
We combine the segmentation masks with the Canny edge \cite{canny1986computational} version of the real images --- see Section \ref{subsec:combined_conditionality} ---, and experiment with various types of prompts --- see Section \ref{subsec:prompt} --- to identify the configuration that leads to the most realistic results. Our approach is therefore twofold, considering on one side the image conditionality, and on another side prompt conditionality. 
An illustration depicting the overview of our approach is presented in Fig. \ref{fig:apprach}. 


Starting from the RailSem19 dataset, the generated images will be used then to increase the size of the dataset and improve Semantic Segmentation performance.





\subsection{Combined Conditionality}
\label{subsec:combined_conditionality}

\begin{figure}[hbt!]
\centering
\centerline{\includegraphics[scale=0.18]{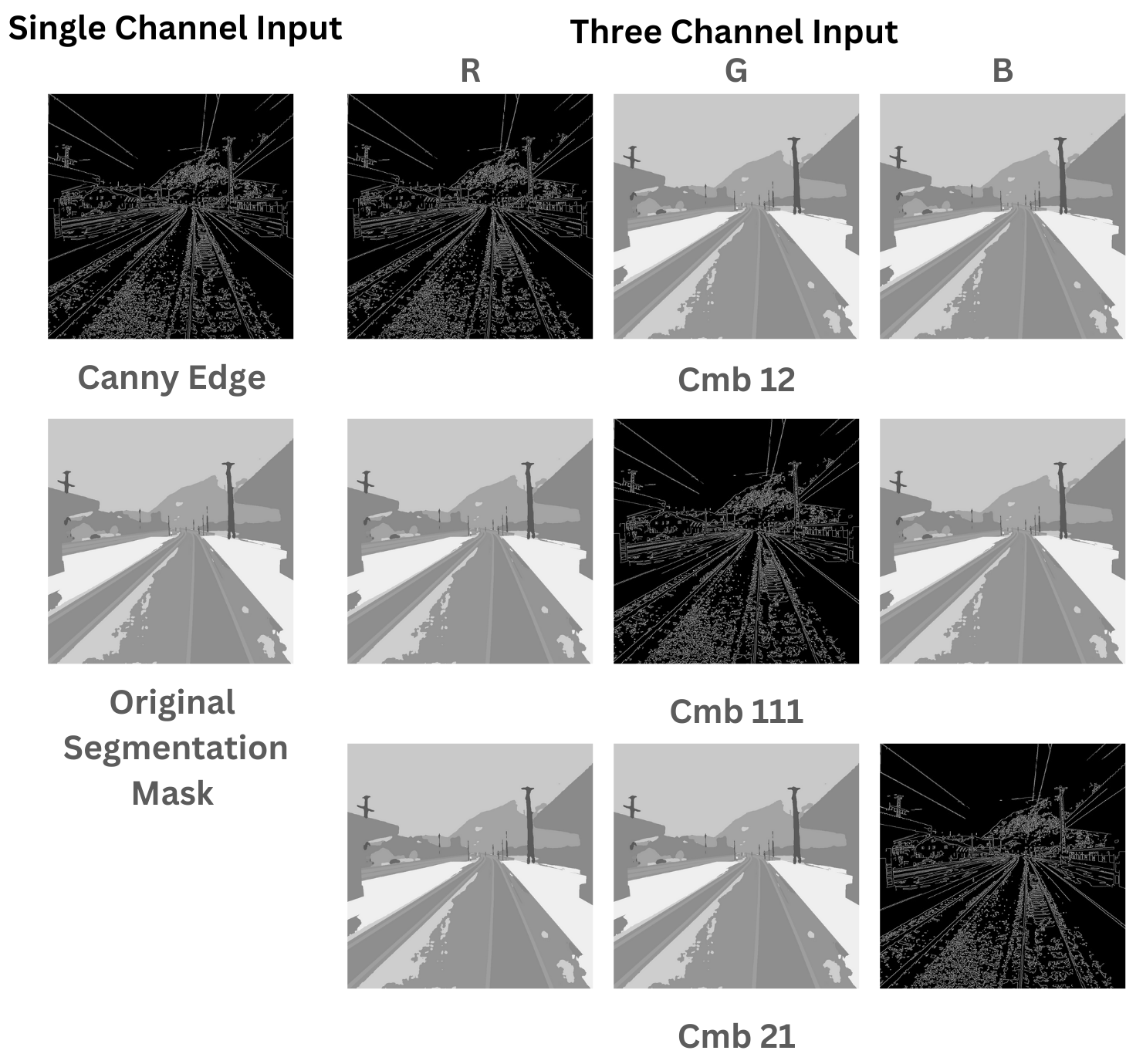}}
	\caption{Conditional representations used as input.}
	\label{fig:conditional_representation}
\end{figure}

ControlNet allows merging multiple conditional inputs during inference time, after each individual model was already trained by considering one condition only. Through a combined representation of more conditions, we enable the network to jointly exploit the complementary information of multiple conditions during the fine-tuning phase.

We fine-tune the Stable Diffusion model used by ControlNet in order to embed additional railway information in the model. By default, the Stable Diffusion model takes as input RGB images, however our default conditional segmentation mask from RailSem19 is represented as a grayscale image. Thus we extend the original grayscale mask by duplicating it for each of the three input channels. The same applies when using the Canny edge image since this type of input condition also occupies a single channel. 

After some initial experiments in which we attempted to change the native number of input channels in order to use only one channel (two channels for the combined representation), we decided to continue with the default number of three channels as this approach yielded much better results. Therefore, to obtain a combined representation of two conditions we assign the original segmentation mask to two of the channels while assigning the Canny edge mask to another channel. Depending on which of the channels received the Canny edge mask, we obtain three possible inputs: Cmb12 (Edge, Mask, Mask), Cmb111 (Mask, Edge, Mask) and Cmb21 (Mask, Edge, Edge), as seen in Fig. \ref{fig:conditional_representation}. 




This novel approach offers two significant advantages:

1. Single Model Training: Regardless of the number of conditions to be considered, only a single model needs to be trained. This simplifies the training process and reduces computational resources.

2. Guided Diffusion Process: The combined condition representation guides the fine-tuning process step by step, simultaneously taking into account all conditions. This ensures a more coherent and integrated output, leveraging the full spectrum of available information.

In contrast, the original version of ControlNet requires multiple models to be trained, each corresponding to a different condition. This renders the proposed method more efficient and effective.

\subsection{Prompting}
\label{subsec:prompt}


The Stable Diffusion model behind ControlNet relies on textual information which describes the desired generated output. In order to obtain a prompt for all images from a dataset, a zero-shot image-to-text generator can be used (e.g. CLIP \cite{radford2021learning}, ZeroCap \cite{tewel2022zerocap}, BLIP-2 \cite{li2023blip}).
In our approach, we actually use the BLIP-2 model which seems to accurately describe the images from our setting. BLIP-2 achieves state-of-the-Art performance on various vision-language tasks \cite{li2023blip}, being efficient and versatile.

Moreover, as recommended by the ControlNet authors, we add a constant decorator to all prompts to enhance image quality: "high quality, extremely detailed, 4K".

We also experiment with multiple types of prompts. In our first experiments we use no textual prompt (\emph{No pr.}). Then we use the same prompt for all images (\emph{Fixed pr.}). Lastly, the most promising technique from our experimental setup was reached through the use of generated prompts, obtained via BLIP-2 (\emph{BLIP-2 pr}).

On top of the three configurations, we also consider negative prompts (\emph{Neg. pr.}) \cite{liu2022compositional} which can be used to guide the image generation model away from common mistakes. This type of prompt can contain text like "low quality-image", "bad anatomy", "unrealistic rails" \cite{ahfaz2024negative}.

\section{Numerical experiments}
\label{sec:results}

In this section, we report the results obtained, provide details about the dataset, and present the selected evaluation means. We present the numerical results of our experiments aimed at evaluating the impact of synthetic images, generated using the proposed ContRail, on the performance of a segmentation model. Our investigation focuses on two key aspects: firstly, how varying parameters within the ControlNet architecture influence the image generation process, and secondly, the extent to which these generated images, when combined with real images, enhance the training and overall performance of the segmentation model. Through detailed analysis and comparison, we aim to demonstrate the potential benefits and practical applications of integrating synthetic data in segmentation tasks.

\subsection{Datasets}
\label{subsec:datasets}

The images used for training and validating the model are sourced from RailSem19 \cite{zendel2019railsem19}, the most comprehensive Semantic Segmentation dataset for railway scenes. It contains 8500 high-quality images obtained from the ego-view of moving trains. The dataset is robust as it showcases scenes with different weather and lighting conditions, and different types of rails, for train and city trams.

Similar to \cite{alexandrescu2022dynamic}, we use a cropped version of the dataset that resizes the images from 1920x1080 to 1080x1080. This aims to reshape the images to a size that fits the aspect ratio of the input layer of the network, without skewing the objects. While the left and right edges of the images are cut to obtain the new size, they do not contain essential information. The rails of interest are located at the center of the image.

\subsection{Evaluation Metrics}
\label{subsec:evaluationMetrics}

To measure the quality of the generated images, we employ both qualitative and quantitative evaluations. The qualitative evaluation involves visually inspecting a limited number of generated samples. There are key aspects we consider quintessential when attempting to replicate the real distribution of railway scene images, the most important being rail quality. In our early experiments, we observed a tendency for the model to produce scenes with two main anomalies: doubled and splintered rails. Through weight improvements and more epochs, the quality of the generated rails and surrounding scenes reached a more than satisfactory state, reducing the occurrence of such cases.

To evaluate numerous sets of synthetic images we use Fréchet Inception Distance (FID) \cite{heusel2017gans} as our quantitative metric. FID measures the Fréchet distance between the distribution of real and generated images. These distributions are described by the mean and variance of the images, represented as 2048 features obtained by passing them through the InceptionNetV3 network \cite{szegedy2016rethinking}. Being a distance metric, a lower FID value suggests an increased image quality.

Note however that, for safety-critical applications, FID is not a reliable score. InceptionNetV3 is a biased estimator which may provide non-deterministic embeddings for different trainings. This means that Inception features are not normally distributed and evaluation on real-world data may not reflect the true potential of the trained model. Despite this, we consider it to quantitatively evaluate our results in this article.


Besides evaluating the generated images in a vacuum either qualitatively or quantitatively, another important test is to measure the effect of the synthetic images for a model tasked with identifying aspects from real images. In our test, we train a segmentation model using both real and generated samples, while evaluating it only on real images. In these experiments we use Intersection-over-Union (IoU) also known as the Jaccard metric \cite{jaccard1912distribution}, due to its ubiquitous character for the task of semantic segmentation.

\subsection{Image Generation}
\label{subsec:resultsImageGeneration}

For the image generation experiments, we consider two dimensions with variable parameters: condition type and prompt type. For types of conditioning images, we choose to work with single-conditioned input images, and multi-conditioned input images. Single-conditioned are the grayscale versions of the original segmentation masks from RailSem19, and the Canny edges representation extracted from the real images by the classic Canny algorithm \cite{canny1986computational}. 
Multi-conditioned images are obtained by combining the grayscale segmentation mask with the Canny edges on specific channels: 12, 21 and 111, as explained in Subsection \ref{subsec:combined_conditionality}.
For types of prompts, we consider: no prompt, a fixed prompt, a prompt generated with BLIP-2, and a negative prompt. In case of all prompts, we concatenate a string to boost quality of results: "high quality, extremely detailed, 4K, HQ" \cite{zhang2023adding}. It is a common prompt engineering technique to add such keywords in the query prompt.

We consider all 8500 images from the RailSem dataset and split them into 6800 training and 1700 validation images, which constitutes an 80:20 split. The images are resized to 512x512. We train ContRail for 13 epochs with a batch size of 4, which results in exactly 22100 training steps. The recommended number of steps to reach the sudden convergence phenomenon is at least 10000 \cite{zhang2023adding}. We run our experiments on NVIDIA GeForce RTX 4090 GPU with 24GB of VRAM on 56GB of RAM. One experiment takes around 7 hours.

We report the FID results between the validation set and the generated images in Table \ref{tab:fid_scores}. Each cell in the table corresponds to one configuration executed once.
The scores are overall varied. The best score is registered by combining the segmentation mask and the Canny representation in Cmb111 without prompts. When using only the segmentation mask condition, the best scores are obtained with BLIP2 prompts. Negative prompts influence the results two-fold: FID scores worsen in combined conditions, while they remain constant without combining conditions.

\begin{table}[h]
    \centering
    \scriptsize
    \begin{tabular}{|l|c|c|c|c|c|}
        \hline
        Prompts $ \backslash $ Cond. & Seg. masks & Canny & Cmb12 & Cmb21 & Cmb111 \\
        \hline
        \textit{No pr.} & 22.92 & 27.67 & \textit{16.57} & 19.65 & \textbf{\textit{16.50}} \\
        \hline
        \textit{Fixed pr.} & 23.80 & 24.22 & 17.46 & 19.65 & \textbf{17.43} \\
        \hline
        \textit{BLIP-2 pr.} & \textit{20.50} & \textit{22.28} & 17.29 & \textit{17.47} & \textbf{17.21} \\
        \hline
        \textit{Neg. pr.} & \textbf{22.24} & 26.34 & 23.62 & 23.79 & 23.53 \\
        \hline
        \textit{Fixed pr. + Neg. pr.} & \textbf{22.53} & 25.55 & 23.74 & 24.04 & 23.46 \\
        \hline
        \textit{BLIP-2 pr. + Neg. pr.} & \textbf{20.78} & 24.14 & 22.28 & 23.63 & 22.22 \\
        \hline
    \end{tabular}
    \caption{FID Scores on various configurations. We marked with bold fonts the best FID score for a particular condition and with italic fonts the best FID score for a particular prompt.}
    \label{tab:fid_scores}
\end{table}

Focusing on the single approach, the scores on the segmentation masks are generally better than those on the Canny representation. The masks provide richer information than simple contours approximated by Canny edges. Many objects resemble similar shapes to rails when represented by edges, such as poles and trees.

\begin{figure}[hbt!]
\centering
\centerline{\includegraphics[scale=0.09]{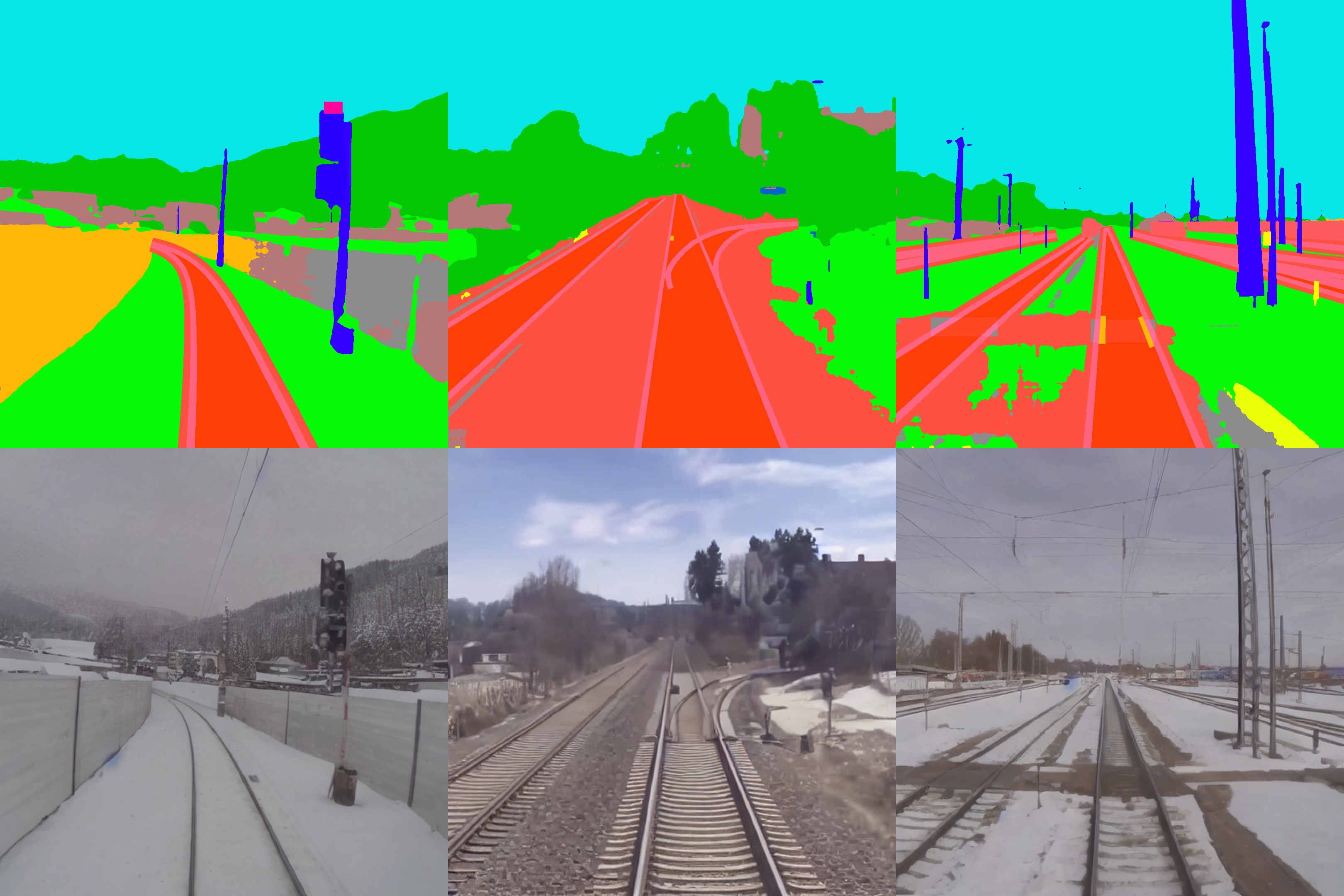}}
	\caption{First row: Segmentation masks; Second row: results obtained using combined masks Cmb111 without training prompts.}
	\label{fig:results_nopr_cmb111}
\end{figure}
\begin{figure}[hbt!]
\centering
\centerline{\includegraphics[scale=0.07]{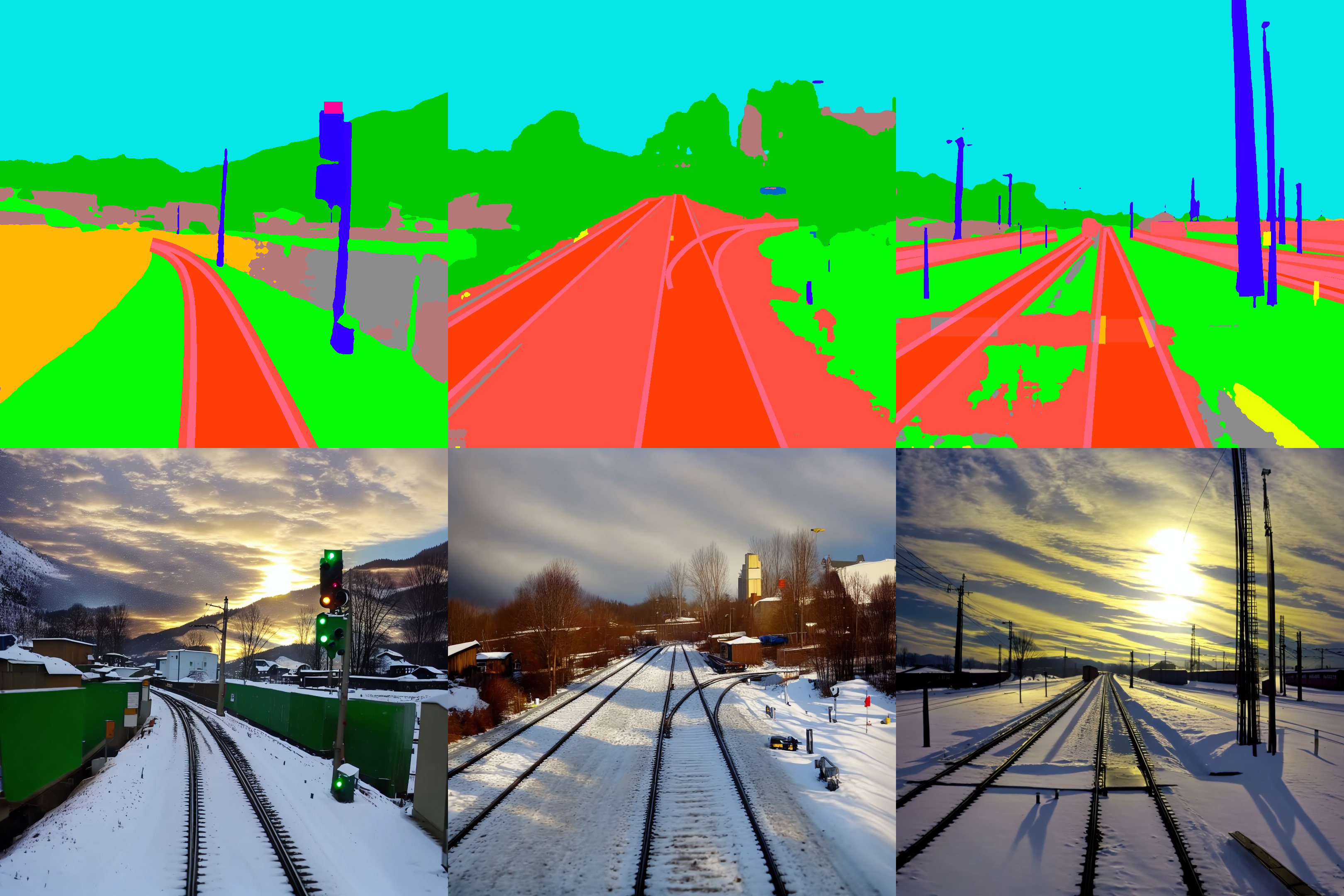}}
	\caption{First row: Segmentation masks; Second row: results obtained using original masks with BLIP-2 prompts.}
	\label{fig:results_BLIP-2_orig_mask}
\end{figure}
\begin{figure}[hbt!]
\centering
\centerline{\includegraphics[scale=0.07]{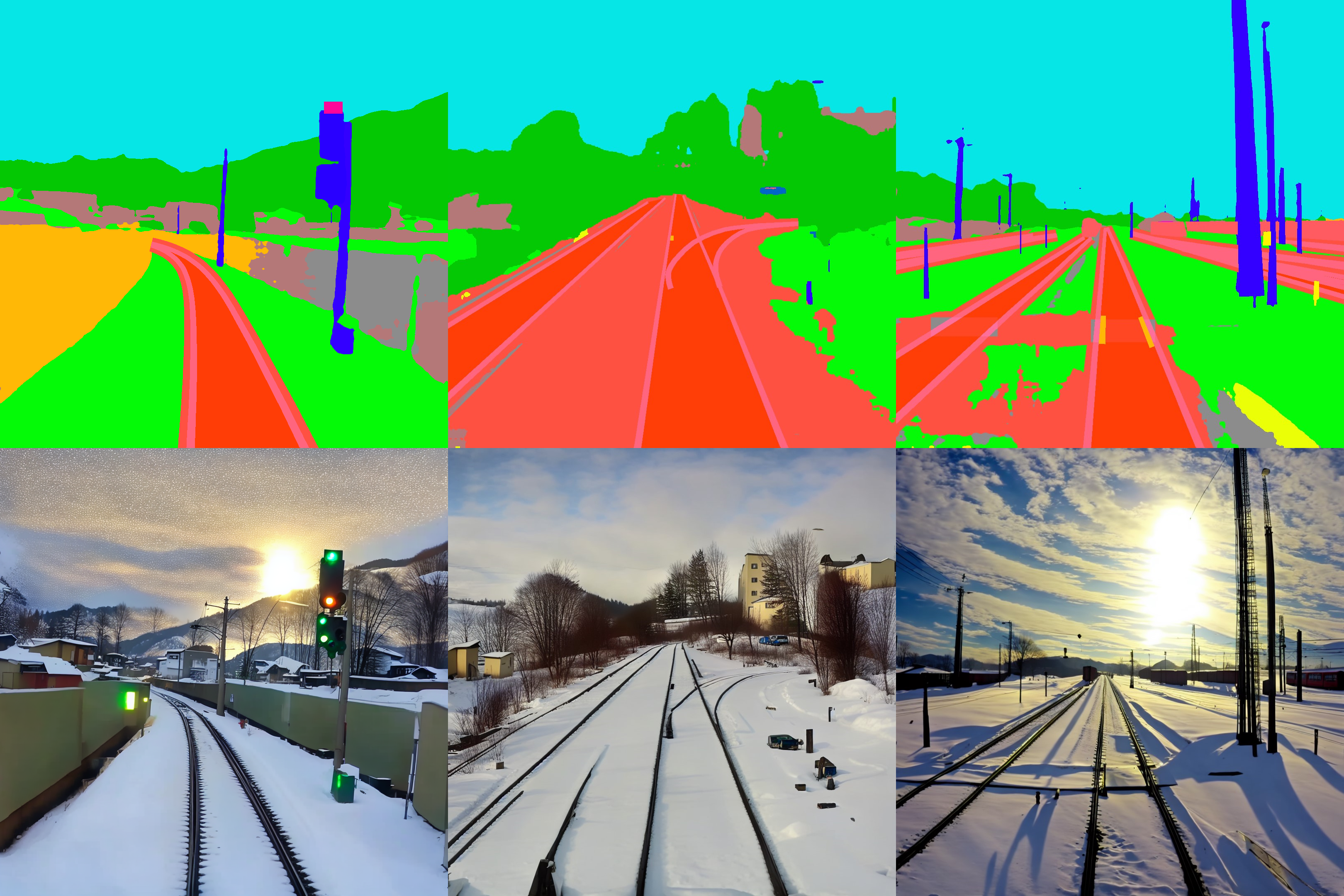}}
	\caption{First row: Segmentation masks; Second row: results obtained using original masks with BLIP-2+Negative prompts.}
	\label{fig:results_BLIP-2_neg_orig_mask}
\end{figure}

We display a selected subset of images for specific configurations: Cmb111 without prompts in Figure \ref{fig:results_nopr_cmb111} (which led to the best quantitative results), original masks with BLIP2 prompts in Figure \ref{fig:results_BLIP-2_orig_mask}, and original masks with BLIP2 prompts and negative prompts in Figure \ref{fig:results_BLIP-2_neg_orig_mask}.

While from the FID perspective the lowest score is obtained on Cmb111 without prompts, visually, the most realistic high-quality images are obtained when using the original masks. This is caused by some low quality, cluttered or complex light scenarios from RailSem19. Some of them are blurred, darker or in a lower resolution. Thus, we consider that a lower-quality image style yields a better score than a high-quality image with a realistic representation that is further away from the mean of the validation data distribution.

We observe the difference in image complexity when comparing the Cmb111 results with the original masks results. In Figure \ref{fig:results_nopr_cmb111}, the synthetic images are less detailed with non essential information such as shadows or lightning effects. They are more focused on a realistic railway scenario, mainly on generating rails, rail track and switches correctly.

In Figure \ref{fig:results_BLIP-2_orig_mask}, note the continuity of the rails, the shadows, and the switch in the middle image. However, the traffic light in the first image is not as realistic as expected, likely due to the fewer samples available for that class. Adding negative prompts in Figure \ref{fig:results_BLIP-2_neg_orig_mask} shows no clear improvements, except for the less realistic traffic signs.

\subsection{Semantic Segmentation results}
\label{subsec:resultsSemSeg}

To gain a better understanding of the use of synthetic images, we consider a set of experiments where synthetic images are used for training, either to replace real images, or to support them. We formulate the semantic segmentation problem as a binary problem, highlighting only the rail class \cite{alexandrescu2022dynamic}. The images are cropped to 1080x1080, resulting in a rail:background ratio of 1:37 \cite{alexandrescu2022dynamic}.

The segmentation model we train is the U-Net \cite{ronneberger2015u} neural network architecture. This architecture is composed of two parts: a feature extractor encoder and a decoder that maps the features back to a segmentation map. The network is suitable for high-resolution semantic segmentation tasks, featuring skip-connections between the encoder and the decoder to ensure information and location flow. The competitive results obtained in \cite{alexandrescu2022dynamic} inspired the network choice.

We train the network for 40 epochs using the Adam optimizer with a learning rate of $1e^{-4}$ and batch size 4 on images of size 512x512. The synthetic images are generated with a single condition (the segmentation masks) and a prompt generated by BLIP-2; this configuration led to the best results visually.

Multiple segmentation models are built. In each setup, the same 500 original real images are considered for validation. Training images are taken in the order they are indexed in the original dataset \cite{zendel2019railsem19}. We further describe the configuration we chose for each segmentation model. Every configuration trains a model from scratch. Synthetic images are all generated using the previously mentioned configuration.

We train the segmentation model on the first 3000 real images from RailSem19 (A). Then, we replace the 3000 real images with their synthetic generated version and train a new network (B).

We then join the real images from (A) with the synthetic images from (B), obtaining a training corpus of 6000 images (C). Note that there are only 3000 unique masks since each is associated with a real and a synthetic image.

The next setup considers a subset of the 3000 images from the previous setups: 1500 real images and 1500 synthetic images, thus gathering a mixed set of 3000 images (D). Similar to (C), the real and synthetic images are associated to the same ground-truth masks.

In the final setups E and F, we again consider 1500 real and 1500 synthetic images, however, this time corresponding to different ground-truth masks: the first 1500 are real, the following 1500 are synthetic (E), and the first 1500 are synthetic, the next 1500 are real (F). 



Each experiment is repeated three times. The mean IoU and standard deviation results on a scale from 0 to 100 are displayed in Table \ref{tab:sem_seg_results}, along with the real and synthetic dataset counts. We restate that the same 500 validation images are considered for each experiment.

\begin{table}[h]
    \centering
    \scalebox{0.75}{
    \begin{tabular}{|l|c|c|c|c|}
        \hline
        Setup & \#Real & \#Synthetic & mIoU & Standard Deviation \\
        \hline
        A & 3000 & 0 & 78.493 & 0.293 \\
        \hline
        B & 0 & 3000 & 78.295 & 0.546 \\
        \hline
        C & 3000 & 3000 & 80.208 & 0.542 \\
        \hline
        D & 1500 & 1500 & 78.369 & 0.253 \\
        \hline
        E & 1500 & 1500 & 77.453 & 0.383 \\
        \hline
        F & 1500 & 1500 & 75.604 & 0.445 \\
        \hline
    \end{tabular}
    }
    \caption{Rail Semantic Segmentation results comparison on three different seeds.}
    \label{tab:sem_seg_results}
\end{table}

Notice how similar results for setups A and B are due to the same number of training images. The standard deviation is slightly higher in setup B when synthetic images are used. Besides that, replacing real images with synthetic ones does not lead to major performance discrepancies.

In setup C we consider twice as many images, resulting in a higher mIoU performance. We may affirm that the usage of a larger number of synthetic images does help the overall performance. This is especially useful when the training dataset is small. In this case, we enrich the corpus of real images with their synthetic counterpart and reuse the same masks to replicate a real scenario. 

Next, in setup D we consider 1500 real images and 1500 synthetic images obtained on the same masks as the real ones. This is similar to setup (C), but with 1500 images instead of 3000. The semantic segmentation model performance does not improve significantly in comparison to setups (A) and (B). The same number of images is used, the only difference being that in setup D every mask is used twice, leading to a less diverse training set.

In setup E we tried to use 3000 unique masks, 1500 for real and 1500 for synthetic images. This led to worse results, probably due to the diverse nature of the data. Training the model on a set of 1500 unique masks twice each epoch leads to better scores than training on 3000 unique masks once each epoch. When inverting the real and synthetic sets of data in setup F, we notice even worse results, thus reaching the same conclusion as before. We attribute these results to the discrepancies between real and synthetic data.


\subsection{Discussion}
\label{subsec:analysis}

To enhance a generative model's ability to create realistic images, it is essential to minimize the difference between real and synthetic data distributions. Training on synthetic images that do not resemble real scenarios can decrease performance and reduce the predicting model's ability to recognize them.

Thus, realistic images are required, however what humans perceive to be realistic may not be perceived similarly by the models. Slight differences such as railway width, traffic light aspect ratio or light effects may negatively influence the learnt distribution and reduce the model's ability to recognize new scenarios. 

Therefore, the image generation models considered should be trained on realistic and representative images. This leads to an infinite loop, since we reach the initial point: the lack of data to train large models. 

This is one common dilemma regarding the issue: if enough data was available to train the image generation models to perform perfect at their task, then why is there a need for new realistic images to be generated? There are many possible answers, however the most resembling one is in case of specific rare scenarios that are hard to capture in reality. Generative models are able to patch together such scenarios and generate realistic images with rare scenarios, which in turn increase the robustness of task-specific models trained on the synthetic data.

\section{Conclusions and Future Considerations}
\label{sec:conclusion}

In this work, we studied the task of Railway Image Generation by employing additional control to the generated images using ControlNets. Our quantitative and qualitative evaluations show small FID scores that indicate a high realism in our images. We also showed how synthetic images enhance the performance of a task specific model, i.e. Rail Semantic Segmentation, using synthetic generated images.

One significant advantage of this approach compared to the results presented in Section \ref{sec:relWork} is the ability to generate images in any context, by using the knowledge acquired by the Stable Diffusion model \cite{runway2022stable} during its training. Most data augmentations using generative models trained from scratch face the issue of producing data from the same distribution as the training set, which may not necessarily add substantial new information. While this type of generation is useful in the medical field, where it is less likely to face drastic changes from the existent data in practice, it can be challenging in the rails domain, because the main issue is caused by the lack of datasets covering a wide range of scenarios, especially rare ones, due to the difficulty and cost involved in capturing those. However, with the help of ControlNet \cite{zhang2023adding}, this problem is mitigated, because this model can generate data in various other scenarios than the ones present in its training set, and the model trained on the new data achieves improved performance, as seen in Table \ref{tab:sem_seg_results}.

Some aspects that can be improved in the future are: obtain more granular segmentation masks to train on more accurate mappings; test our approach on object detection and assess its performance using synthetic data; experiment with more input types (Holistically-nested Edges \cite{xie2015holistically}, Multi-Line Segment Detection).


{\small
\balance
\bibliographystyle{ieee_fullname}

\begin{thebibliography}{99}
  \bibitem{ahfaz2024negative} 
  Ahfaz Ahmed. A Huge List of Stable Diffusion Negative Prompts. \url{https://huggingface.co/runwayml/stable-diffusion-v1-5}, 2024.

  \bibitem{alexandrescu2022dynamic}
  Andrei-Robert Alexandrescu and Alexandru Manole. A dynamic approach for railway semantic segmentation. \textit{Studia Universitatis Babes-Bolyai Informatica}, pages 61–76, 2022.

  \bibitem{bansal2022systematic}
  Ms Aayushi Bansal, Dr Rewa Sharma, and Dr Mamta Kathuria. A systematic review on data scarcity problem in deep learning: solution and applications. \textit{ACM Computing Surveys (Csur)}, 54(10s):1–29, 2022.

  \bibitem{canny1986computational}
  John Canny. A computational approach to edge detection. \textit{IEEE Transactions on Pattern Analysis and Machine Intelligence}, 8(6):679–698, 1986.

  \bibitem{carlson2023generation}
  Johanna Carlson and Lovisa Byman. Generation of synthetic traffic sign images using diffusion models. Dissertation, 2023.

  \bibitem{dice1945measures}
  Lee R Dice. Measures of the amount of ecologic association between species. \textit{Ecology}, 26(3):297–302, 1945.

  \bibitem{eigenschink2023deep}
  Peter Eigenschink, Thomas Reutterer, Stefan Vamosi, Ralf Vamosi, Chang Sun, and Klaudius Kalcher. Deep generative models for synthetic data: A survey. \textit{IEEE Access}, 11:47304–47320, 2023.

  \bibitem{fernandez2022can}
  Virginia Fernandez, Walter Hugo Lopez Pinaya, Pedro Borges, Petru-Daniel Tudosiu, Mark S Graham, Tom Vercauteren, and M Jorge Cardoso. Can segmentation models be trained with fully synthetically generated data? In \textit{International Workshop on Simulation and Synthesis in Medical Imaging}, pages 79–90. Springer, 2022.

  \bibitem{frid2018gan}
  Maayan Frid-Adar, Idit Diamant, Eyal Klang, Michal Amitai, Jacob Goldberger, and Hayit Greenspan. GAN-based synthetic medical image augmentation for increased CNN performance in liver lesion classification. \textit{Neurocomputing}, 321:321–331, 2018.

  \bibitem{goodfellow2014generative}
  Ian Goodfellow, Jean Pouget-Abadie, Mehdi Mirza, Bing Xu, David Warde-Farley, Sherjil Ozair, Aaron Courville, and Yoshua Bengio. Generative adversarial nets. In \textit{Advances in Neural Information Processing Systems}, 27, 2014.

  \bibitem{ha2016hypernetworks}
  David Ha, Andrew Dai, and Quoc V Le. Hypernetworks. \textit{arXiv preprint arXiv:1609.09106}, 2016.

  \bibitem{heusel2017gans}
  Martin Heusel, Hubert Ramsauer, Thomas Unterthiner, Bernhard Nessler, and Sepp Hochreiter. GANs trained by a two time-scale update rule converge to a local Nash equilibrium. In \textit{Advances in Neural Information Processing Systems}, 30, 2017.

  \bibitem{jaccard1912distribution}
  Paul Jaccard. The distribution of the flora in the alpine zone. \textit{New Phytologist}, 11(2):37–50, 1912.

  \bibitem{jiang2024scedit}
  Zeyinzi Jiang, Chaojie Mao, Yulin Pan, Zhen Han, and Jingfeng Zhang. SCEDIT: Efficient and controllable image diffusion generation via skip connection editing. In \textit{Proceedings of the IEEE/CVF Conference on Computer Vision and Pattern Recognition}, pages 8995–9004, 2024.

  \bibitem{johnson2016perceptual}
  Justin Johnson, Alexandre Alahi, and Li Fei-Fei. Perceptual losses for real-time style transfer and super-resolution. In \textit{Computer Vision–ECCV 2016: 14th European Conference}, Amsterdam, The Netherlands, October 11-14, 2016, Proceedings, Part II 14, pages 694–711. Springer, 2016.

  \bibitem{kebaili2023deep}
  Aghiles Kebaili, Jérôme Lapuyade-Lahorgue, and Su Ruan. Deep learning approaches for data augmentation in medical imaging: a review. \textit{Journal of Imaging}, 9(4):81, 2023.

  \bibitem{khullar2023synthetic}
  Dipika Khullar, Yash Shah, Ninad Kulkarni, and Negin Sokhandan. Synthetic data generation for scarce road scene detection scenarios. In \textit{NeurIPS 2023 Workshop on Synthetic Data Generation with Generative AI}, 2023.

  \bibitem{li2023blip}
  Junnan Li, Dongxu Li, Silvio Savarese, and Steven Hoi. BLIP-2: Bootstrapping language-image pre-training with frozen image encoders and large language models. In \textit{International Conference on Machine Learning}, pages 19730–19742. PMLR, 2023.

  \bibitem{liu2022compositional}
  Nan Liu, Shuang Li, Yilun Du, Antonio Torralba, and Joshua B Tenenbaum. Compositional visual generation with composable diffusion models. In \textit{European Conference on Computer Vision}, pages 423–439. Springer, 2022.

  \bibitem{mou2024t2i}
  Chong Mou, Xintao Wang, Liangbin Xie, Yanze Wu, Jian Zhang, Zhonggang Qi, and Ying Shan. T2I-adapter: Learning adapters to dig out more controllable ability for text-to-image diffusion models. In \textit{Proceedings of the AAAI Conference on Artificial Intelligence}, volume 38, pages 4296–4304, 2024.

  \bibitem{park2019semantic}
  Taesung Park, Ming-Yu Liu, Ting-Chun Wang, and Jun-Yan Zhu. Semantic image synthesis with spatially-adaptive normalization. In \textit{Proceedings of the IEEE/CVF Conference on Computer Vision and Pattern Recognition}, pages 2337–2346, 2019.

  \bibitem{perera2023analyzing}
  Malsha V Perera and Vishal M Patel. Analyzing bias in diffusion-based face generation models. In \textit{2023 IEEE International Joint Conference on Biometrics (IJCB)}, pages 1–10. IEEE, 2023.

  \bibitem{radford2021learning}
  Alec Radford, Jong Wook Kim, Chris Hallacy, Aditya Ramesh, Gabriel Goh, Sandhini Agarwal, Girish Sastry, Amanda Askell, Pamela Mishkin, Jack Clark, et al. Learning transferable visual models from natural language supervision. In \textit{International Conference on Machine Learning}, pages 8748–8763. PMLR, 2021.

  \bibitem{rombach2022high}
  Robin Rombach, Andreas Blattmann, Dominik Lorenz, Patrick Esser, and Björn Ommer. High-resolution image synthesis with latent diffusion models. In \textit{Proceedings of the IEEE/CVF Conference on Computer Vision and Pattern Recognition}, pages 10684–10695, 2022.

  \bibitem{ronneberger2015u}
  Olaf Ronneberger, Philipp Fischer, and Thomas Brox. U-Net: Convolutional networks for biomedical image segmentation. In \textit{Medical Image Computing and Computer-Assisted Intervention–MICCAI 2015: 18th International Conference}, Munich, Germany, October 5-9, 2015, proceedings, part III 18, pages 234–241. Springer, 2015.

  \bibitem{runway2022stable}
  Runway. Stable Diffusion v1-5 Model Card. \url{https://huggingface.co/runwayml/stable-diffusion-v1-5}, 2024.

  \bibitem{schuhmann2022laion}
  Christoph Schuhmann, Romain Beaumont, Richard Vencu, Cade Gordon, Ross Wightman, Mehdi Cherti, Theo Coombes, Aarush Katta, Clayton Mullis, Mitchell Wortsman, et al. LAION-5B: An open large-scale dataset for training next generation image-text models. \textit{Advances in Neural Information Processing Systems}, 35:25278–25294, 2022.

  \bibitem{sohl2015deep}
  Jascha Sohl-Dickstein, Eric Weiss, Niru Maheswaranathan, and Surya Ganguli. Deep unsupervised learning using nonequilibrium thermodynamics. In \textit{International Conference on Machine Learning}, pages 2256–2265. PMLR, 2015.

  \bibitem{szegedy2016rethinking}
  Christian Szegedy, Vincent Vanhoucke, Sergey Ioffe, Jon Shlens, and Zbigniew Wojna. Rethinking the inception architecture for computer vision. In \textit{Proceedings of the IEEE Conference on Computer Vision and Pattern Recognition}, pages 2818–2826, 2016.

  \bibitem{tewel2022zerocap}
  Yoad Tewel, Yoav Shalev, Idan Schwartz, and Lior Wolf. ZeroCap: Zero-shot image-to-text generation for visual-semantic arithmetic. In \textit{Proceedings of the IEEE/CVF Conference on Computer Vision and Pattern Recognition}, pages 17918–17928, 2022.

  \bibitem{whang2023data}
  Steven Euijong Whang, Yuji Roh, Hwanjun Song, and Jae-Gil Lee. Data collection and quality challenges in deep learning: A data-centric AI perspective. \textit{The VLDB Journal}, 32(4):791–813, 2023.

  \bibitem{xie2015holistically}
  Saining Xie and Zhuowen Tu. Holistically-nested edge detection. In \textit{Proceedings of the IEEE International Conference on Computer Vision}, pages 1395–1403, 2015.

  \bibitem{zendel2019railsem19}
  Oliver Zendel, Markus Murschitz, Marcel Zeilinger, Daniel Steininger, Sara Abbasi, and Csaba Beleznai. Railsem19: A dataset for semantic rail scene understanding. In \textit{Proceedings of the IEEE/CVF Conference on Computer Vision and Pattern Recognition Workshops}, pages 0–0, 2019.

  \bibitem{zhang2023adding}
  Lvmin Zhang, Anyi Rao, and Maneesh Agrawala. Adding conditional control to text-to-image diffusion models. In \textit{Proceedings of the IEEE/CVF International Conference on Computer Vision}, pages 3836–3847, 2023. 
\end{thebibliography}

\end{document}